\documentclass{vldb}

\usepackage{amsmath}
\usepackage{graphicx}
\usepackage{balance}  
\usepackage{array}
\usepackage{multirow}
\usepackage{subfigure}
\usepackage[colorinlistoftodos]{todonotes}
\usepackage[colorlinks=true, allcolors=blue]{hyperref}
\newcommand{\SM}[1]{{\color{red}[SM: #1]}}

\newcommand{\vek}[1]{\ensuremath{\text{\bf {#1}}}}
\newcommand{\vx}{{\vek{x}}}
\newcommand{\vy}{{\vek{y}}}
\newcommand{\vs}{{\vek{s}}}
\newcommand{\vh}{{\vek{h}}}
\def\etal{\emph{et al}.}

\vldbTitle{AR-MDN: Associative and Recurrent Mixture Density Networks for eRetail Demand Forecasting}
\vldbAuthors{Srayanta Mukherjee, Devashish Shankar, Atin Ghosh, Nilam Tatha-wadekar, Pramod Kompalli, Sunita Sarawagi, Krishnendu Chaudhury}
\vldbDOI{https://doi.org/TBD}

\begin{document}

\title{AR-MDN: Associative and Recurrent Mixture Density Networks for eRetail Demand Forecasting}
\author{Srayanta Mukherjee, Devashish Shankar, Atin Ghosh, Nilam Tathawadekar, \\ Pramod Kompalli, Sunita Sarawagi{$^\dagger$},  Krishnendu Chaudhury  \\ Flipkart.com, Bengaluru, India. \\  {$^\dagger$}IIT-Bombay, Mumbai, India \\email: srayanta@gmail.com}

\maketitle

\begin{abstract}

Accurate demand forecasts can help on-line retail organizations better plan their supply-chain processes. The challenge, however, is the large number of associative factors that result in large, non-stationary shifts in demand, which traditional time series and regression approaches fail to model. In this paper, we propose a Neural Network architecture called AR-MDN, that simultaneously models associative factors, time-series trends and the variance in the demand. We first identify several causal features and use a combination of feature embeddings, MLP and LSTM to represent them. We then model the output density as a learned mixture of Gaussian distributions. The AR-MDN can be trained end-to-end without the need for additional supervision. We experiment on a dataset of an year's worth of data  over tens-of-thousands of products from Flipkart. The proposed architecture yields a significant improvement in forecasting accuracy when compared with existing alternatives.

\end{abstract}

\section{Introduction}
In the retail industry, accurate sales forecasts are essential for timely buying and replenishment of inventory, top-line demand planning and effective supply-chain utilization~\cite{kleinberg03,larson01}. Errors in predicting demand adversely affects multiple business critical metrics such as out-of-stock  percentage, inventory health, wastage of man power and logistical resources.  In this paper we present how we tackle the challenges of demand forecasting for inventory management in Flipkart,  India's largest eRetail company.

In an eRetail setting, demand prediction is particularly challenging, given the huge catalog of products numbering in tens of millions, sale events (such as festival sales), frequent discounting of price, geographically distributed customer base, and competitive interactions 
On the positive side, due to the pervasive nature of eRetail, its huge customer bases, and ``online" mode of transactions,  there is a wealth of data that could be exploited to build accurate models for demand prediction.  For example, Flipkart has an active catalogue of 80 million products,  a daily footfall of 10 million page visits, and sells an average 0.5 million units per day. 


Much of previous demand prediction methods relied on statistical time-series methods such as ARIMA and other state space models that capture parameters like seasonality and trends. However, traditional time-series models are incapable of taking into account a number of associative factors that have strong influence on short term buying decisions, including factors like deep-discounting, bundle offers,  aggressive merchandising campaigns, and competitor trends~\cite{johnson15}.  These factors lead to wild fluctuations on a weekly basis with sharp spikes.  The non-trivial nature of the problem dictates the use of advanced machine learning approaches which are capable of handling the multi-dimensional feature space that drives demand. Our first method of choice  was a Boosted Cubist~\cite{quinlan92} model that is a powerful regression model that combines four learning ideas: a regression tree to capture non-linear feature interaction, a linear regression model at the leaves, a nearest neighbor refinement, and finally a Boosted ensemble.  This, and related models like Gradient Boosted Decision Trees have been found to be highest performing in many prediction challenges\cite{gradientboosting}.


A failing of Cubist~\cite{quinlan92}, and similar regressors, is that it is a scalar prediction model and requires special pre-processing to capture sequential predictions on a time-series.   Also, motivated by the impressive success of modern deep learning methods on various speech, text, and image processing tasks, we next focused on a neural network model for our demand forecasting problem.  However, we soon found that a straight-forward deployment of recurrent neural networks for time series forecasting failed to provide us any gains beyond existing regression models (like Boosted Cubist) on Flipkart's challenging dataset.  In this paper we present the many design and engineering challenges that we had to tackle in implementing a modern deep-learning  model, that substantially out-performed our first generation Boosted Cubist model.


Our multi-layered hybrid deep neural network, called Associative and Recurrent Mixture Density Networks (AR-MDN), can effectively combine both time-series and associative factors. AR-MDN outputs the probability distribution over demands as a mixture of Gaussians, and we show that this is crucial to capture the unpredictable trends in demands in heterogeneous settings such as ours.  We provide engineering solutions for training models along hierarchies of products and geography, and demonstrate the role of our carefully designed feature sets.




\subsection{Problem Setting}
Our point of sales (POS) data contains with each sales transaction a number of properties that potentially influenced that purchase event. These span various properties of the item sold, the geography where sold, and events/properties characterizing the time of sale.   When creating a forecast model the first key decision is choosing a level of granularity of representing the dimensions of item, geography, and time.  

In this work, we focus on demand prediction for inventory management and replenishment, where the demand needs to be estimated at the SKU level. The inventory is typically replenished on a weekly basis, hence, we aim to generate a weekly forecast of demand.  Further, the supply chain can perform efficiently if the customer demands are met from the nearest warehouse.  Therefore, along the geography dimension we assess demand at the warehouse or “fulfillment center” (FC) level. Succinctly, we aim to make predictions at the granularity of SKU $\times$ FC $\times$ Week level along the item, geography, and time dimensions respectively.

\subsection{Contributions}

We highlight the main contributions of this work here.

We present the design of AR-MDN, a deep-learning based forecast model that can grapple with the scale and heterogeneity of the largest eRetail store in India.
Our proposed model incorporates a number of careful design ideas including a mixture of Gaussians at the output layer, a practical training methodology to trade-off sparsity with diversity, and a staged feed-forward, recurrent layer to fuse associative with temporal features.

We compare this network with the best of the breed traditional method based on boosted hybrid trees. We are aware of no earlier such study at the scale of a challenging retail dataset such as ours.

Our experiments on Flipkart's sales data show that a well-engineered deep learning models does indeed live up to their hype of being significantly better than a state of the art non-deep learning based method.  AR-MDN causes between 10 to 25\% reduction in error compared to Boosted Cubist.

We show that our specific neural architecture is significantly better than off-the-shelf deep learning time series models.  Somewhat surprisingly we observe that feature engineering continues to be useful even for deep learning models and we get a 7-20\% improvement in performance by our engineered features.

\subsection{Outline}

The rest of the paper is organised as follows. In Section~\ref{sec:litreview} we present an overview of the extensive literature on forecasting models.  In Section~\ref{sec:armdn} we present AR-MDN, our proposed deep learning based model.  In Section~\ref{sec:expts}  we compare the performance of our proposed model with a state-of-the-art traditional model for time series forecasting, and alternative deep learning models. We conclude and provide future directions in Section~\ref{sec:conc}.

\section{Literature Review}
\label{sec:litreview}
We categorize the massive literature of time series forecasting into four groups and present an overview of each.

\paragraph*{Traditional time series methods}
Traditional time series methods like  Box-Jenkins~\cite{box64,box68} and variations of the state space models like ARIMA  and Holt-Winters\cite{hyndman08}, have been used for demand forecasting in the industry for a very long time. The reason for their popularity is their simplicity. 
While such methods produce admirable results in many cases, for retail demand forecasting, these methods perform poorly. A primary reason for their failure is the presence of several exogeneous casual factors, that make the demand curve highly erratic with sharp peaks and valleys.  Further, the sales time series is often non-stationary and heteroscedastic, thus violating core assumptions of the classical methods. 
Additionally, in large eRetail stores like Flipkart, the product landscape is huge, making the demand curve a function not just of itself but substitutable and complementary products as well. 
To alleviate these problems, a number of innovations have been attempted in recent years that we review next.  

\paragraph*{Wavelet-based approaches} A popular approach to tackle non-stationarity in time-series data, is Wavelet Transforms (WT)~\cite{zhang01, wang02,papagiannnaki03}. The advantage of WT is that it can effectively decompose the time series into its component frequency and time, thereby allowing for identification of primary frequency components. For example, for financial time series prediction, multiple approaches~\cite{bjorn95,moody97} used the WT, modeling the market as fractional Brownian motion. The WT can also be used as a preprocessing to multiple methods such as Artificial Neural Network (ANN)~\cite{zheng99}, Kalman filtering~\cite{cristi00} and an AR model~\cite{soltani00}.
A variant was proposed in Stolojescu~\etal~\cite{stolojescu2010} of using  Stationary WT (SWT) as a preprocessing step to ARIMA, ANN, linear regression and random walk. Overall, the ANN combined with the SWT performed the best among the four models, effectively capturing the non-linearity. They also compared various mother wavelets and concluded that the Haar and Reverse Biorthogonal were most suitable for the task. Zheng~\etal~\cite{zheng99}, Soltani~\etal~\cite{soltani00} and Renauld~\etal~\cite{renaud02} have also explored the use of Haar mother wavelet in time-series forecasting problems. However, wavelet based methods were unsuccessful in producing acceptable results in our case primarily  due to their inability to account for the causal ( associative) features that lead to sharp changes in demand.

\paragraph*{Regression-based methods}
Another approach is to use generic machine learning regressors along with autoregressive features to capture the time-series~\cite{APPELHANS2015}. 
A comparison of the different classes of Regression methods can be found in the work of Wang~\etal~\cite{WANG2016} and Torgo \cite{Torgo1997}. Regression methods rely on minimizing the sum of squared errors or other similar objective functions while achieving a balance between the bias and variance of the model. Particularly robust and high-performing models have traditionally been i) Support Vector Regression (SVRs) that uses non-linear kernel functions 
ii) Random forests, that averages over independent individual decision trees and iii) Gradient boosting which iteratively reduces residual errors using new models. 

Cubist~\cite{quinlan92} is a hybrid of decision trees (or rules) and linear regression models where the predictions are further refined with a nearest neighbor adjustment.
The Cubist regression method has also been found to perform admirably, as compared to other popular regression techniques in forecasting problems as highlighted in the works of Zhang~\etal~\cite{zhang16}, Meyer~\etal~\cite{meyer16} and  Walton~\cite{walton08}. 
However, Cubist and other regression based methods are not designed to explicitly model time-series data. They instead extract intermediate features based on the time-series, which are then treated as independent variables while learning the model. 

\paragraph*{Deep Learning Methods}
Deep Neural Networks have recently been proven to have tremendous potential in modeling tough classification and regression problems.  One of the first work on time-series based forecasting using deep neural networks was by Busseti~\etal~\cite{busseti12} for energy load forecasting. They implemented and compared multiple methods including a deep feed-forward network and a deep recurrent network (RNN), amongst which the RNN was found to perform the best. Similarly Shi~\etal~\cite{shi15} extended a Fully Connected LSTM (FC-LSTM) to have convolution like features (ConvLSTM) for the problem of precipitation forecasting over a localized region for a short time period in the future. Their experiments were able to capture spatio-temporal features better and outperformed more traditional methods. Another deep learning methodology that has been used for time-series forecasting problems are deep belief networks (DBN). Qiu~\etal~\cite{qiu14} applied an ensemble of DBNs aggregated together by a Support Vector Regression (SVR) model to the problem of energy load forecasting. Takashi Kuremoto used multiple Restricted Boltzmann machines (RBM) optimized via Particle Swarm Optimization (PSO) \cite{Kuremoto2014}. Another innovative approach was proposed by Mocanu~\etal~\cite{mocanu16} who used a Factored Conditional RBM (FCRBM) for energy load forecasting. Their method also used additional associative features like electricity price and weather conditions other than the time series only.\\

In retail forecasting, which requires a probabilistic forecasting output, results have mostly been mixed using modern deep-learning techniques~\cite{KOURENTZES2013}. Recently though, Flunkert~\etal~\cite{flunkert17} used a probabilistic LSTM, DeepAR which was both autoregressive and recurrent and was able to demonstrate significant improvements. Key to their success was to learn a global model across the time-series of all items instead of relying on the time-series of a single product alone. This is similar to the strategy that we follow in this work.
Second, they propose to model the output on a Gaussian likelihood for real-valued data and negative binomial likelihood for positive count data. In contrast, our approach is to use a mixture of Gaussians as the output density model.
Importantly, this article holds particular significance for the field as they were able to effectively demonstrate the applicability of modern deep learning techniques on the problem of demand forecasting for eRetail.

\section{Our Prediction Model}
\label{sec:armdn}
In this section we present a deep learning based solution to the challenging task of demand forecasting in large online retail stores.  We start by formally defining the demand forecasting task.  Then in Section~\ref{sec:arch} we describe the architecture of AR-MDN, our proposed forecasting model.  In Section~\ref{sec:train} we describe the training process.


\subsection{Problem Formulation}

\begin{figure}[t]
\begin{center}
  \includegraphics[width=80mm, scale=0.8]{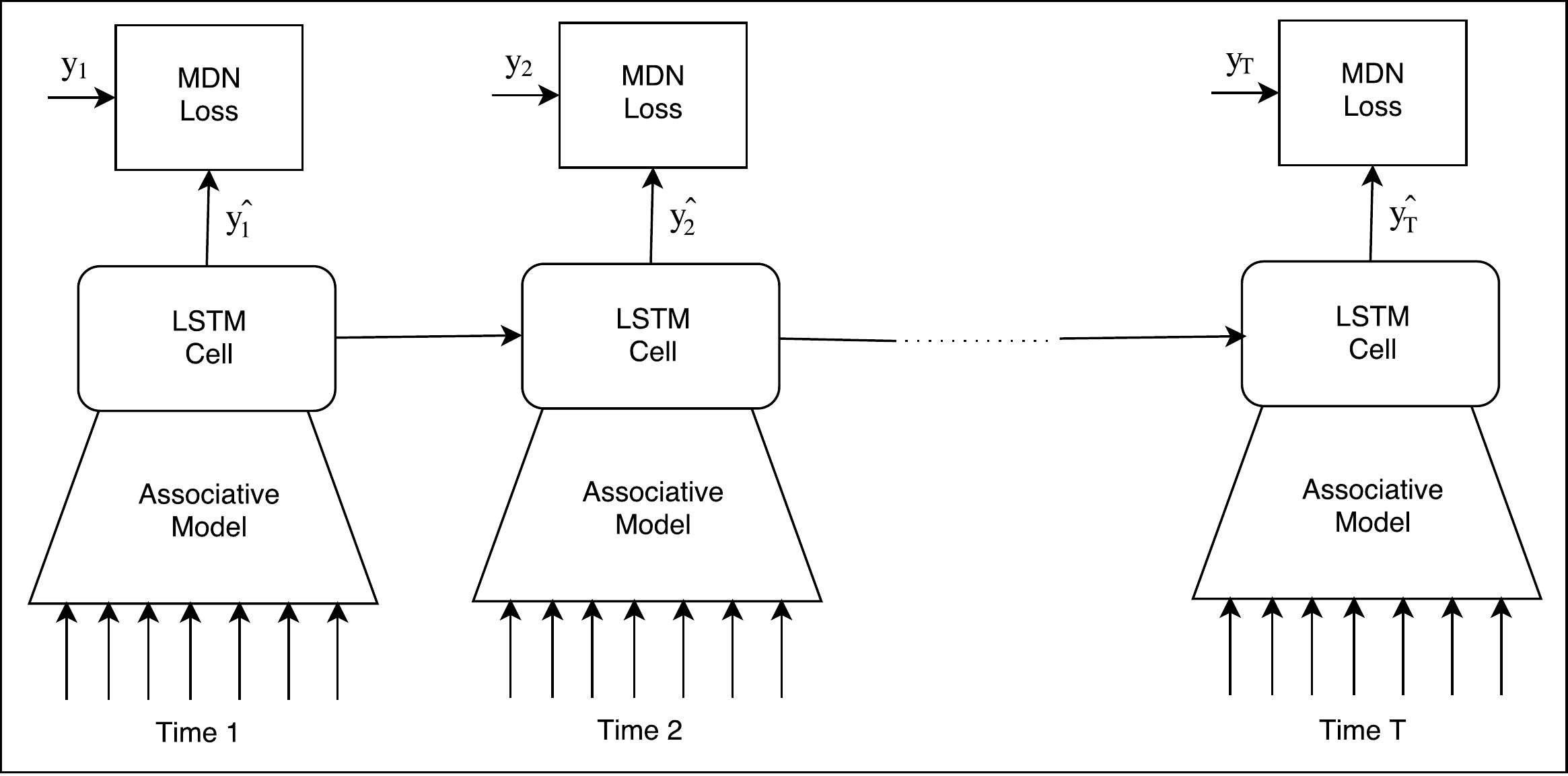}
  \caption{The proposed three-stage architecture of the Associative and Recurrent Mixture Density Networks (AR-MDN). Here we see the model unrolled across time using an LSTM. At each time instant, a set of associative features are modeled by an MLP. The final prediction layer uses an MDN loss. Details are given in Section~\ref{sec:arch}}
  \label{fig:dim1}
\end{center}
\end{figure}

For each SKU ``$s$'' and geographical region ``$r$'', we have a time series of weekly demand values up to a point of time ``$t$'' measured in weeks\footnote{In general, different products might have different lengths of historical data.  We handle these differences in our model but assume for simplicity of writing that all times series are of the same length.}.  Each point in the time series is also associated with a set of causal (or associative) features.  
Some of these features are fixed properties of the product but most of them (including features like price) change along time. We describe them below.

\subsubsection{Features}
\label{sec:features}
Our features capture a large variety of properties of the product, the time, the product's price at the time, advertisements surrounding the sales event, and so on.  
Table~\ref{tab:features} presents  the features we extracted from our point of sales data grouped into five different clusters.  Further, a feature could be numerical, categorical, or binary. 
Some of these are raw features of the data, e.g. the price of the product but we also have a large number of interesting features derived based on our intuitive understanding of consumer behavior.
For example, consider the derived feature called  ``time since last price change''. We observed that when the price of an SKU is decreased from $x$ to $x'$, the demand  increases from $y$ to $y'$ ($y<y'$) for only a brief time period.  Beyond this time period, even if the price is maintained at $x'$, the demand rate decays quickly from $y'$ and stabilizes back to $y$. 
In Section~\ref{sec:derived}, we will show the tremendous impact of these derived features.
 



We use ``$i$''  to index a time series corresponding to a $(s,r)$ combination and denote its feature and demand values along time as $\{(\vx_{i,1}, y_{i,1}),\ldots,(\vx_{i,t}, y_{i,t})\}$ where each $\vx_{i,t}$ denotes the vector of all associative features captured at time $t$ for time series $i$ and $y_{i,t}$ denotes the corresponding demand.  Thus, our available dataset $D$ can be expressed as a set of $N$ time series of the form $$D=\{(s_i,r_i,\vx_{i,1\ldots t}, y_{i,1,\ldots,t}):i=1\ldots N\}$$
$N$ in our setup is very large and is of the order of several million. 

For forecasting, we are given a set of consecutive future time horizons $t+1,\ldots,T$ and the corresponding  input features $\vx_{i,t+1}, \ldots, \vx_{i, T}$ for them. We need to predict the corresponding demand values $\hat{y}_{i,t+1}, \ldots, \hat{y}_{i,T}$.   In order to capture the uncertainty associated with these predictions, we propose to output a distribution instead of a fixed value. 
We use a parametric model that trains parameters $\theta$ to predict such a distribution as
\begin{equation}
P(\vy_{i,t+1 : T}|\vy_{i,1 : t},\vx_{i,1 : T},\theta,s_i,r_i)
\end{equation}

\subsection{The AR-MDN Architecture}
\label{sec:arch}
We propose a new architecture for demand forecasting. The proposed model combines a multi-layer perceptron (MLP) and a long-short term memory (LSTM) network for simultaneous handling of associative and time-series features, and a mixture density network at the output to flexibly capture output uncertainty. The final model, AR-MDN, is therefore associative, recurrent and probabilistic. Figure \ref{fig:dim1} shows a representation of the overall deep net architecture. In what follows, we present detailed descriptions of the three main stages of our model.

\begin{table*}[ht]
\centering
\begin{tabular}{||l|m{6cm}|m{6cm}|l||}
\hline \hline
\bf Demand Factor & \bf Feature Name &  \bf Description &  \bf Data Type \\
\hline
\multirow{ 3}{*}{Product}  & Product ID &  Captures ``latent'' factors of the product & 1-Hot \\
 & Product Tier & Tiers are based on popularity of products & Categorical \\
 & Historical Out-of-Stock \% & Assumed to be always In-Stock during Testing phase & Numerical \\
\hline
\multirow{ 3}{*}{Product Visibility} & Sale event type & Category-level of the Sale event & Categorical \\
 & Deal Card &  & Binary \\
 & Banner on Homepage & & Binary \\
 \hline
\multirow{ 7}{*}{Price} & Listed Price & Govt. mandated Max. Retail Price& Numerical \\
 & Discounted Price & Price offered by Flipkart & Numerical \\
 & Effective Price & Final price after cash-back, product exchange, freebie, etc. & Numerical \\
  & Difference in price from historical mean & Captures ``stable'' price of the product & Numerical \\
  & Historical Min Price  & & Numerical \\
    & Historical Max Price  & & Numerical \\
  & Avg. discount on similar products & Products in the same category are considered similar & Numerical \\
 \hline 
 \multirow{3}{*}{Convenience} & No-cost EMI & & Binary\\
 & Product Exchange & & Binary \\
 & Exclusive to Flipkart & & Binary \\
 \hline 
 
\multirow{ 5}{*}{Time}  & Week of the month & Captures seasonality & Numerical \\
 & Lag feature - Price & Mean of past $n$ weeks' price  & Numerical \\
 & Lag feature - Sale & Mean of past $n$ weeks' sale units  & Numerical \\
 &  Time elapsed since last price change & Sales peak when price changes, but reverts  quickly to previous volume  & Numerical \\
 &  Time since last event or upcoming event & Units sold dips immediately before and after a Sale event & Numerical \\
 & Time  since first order &  Captures ``Newness'' of Product & Numerical \\
\hline \hline
\end{tabular}
\caption{The list of associative features that were identified to  model the demand of a given product. These features are extracted for each time instant ``$t$''. This set includes both raw features that are recorded in the sale data, and derived features that are computed from raw signals. The multi-modality of the signals that need to be modeled for our task, is clear from the data-type column.}
\label{tab:features}
\end{table*}

\begin{figure}[t]
\begin{center}
 \includegraphics[width=80mm, scale=0.7]{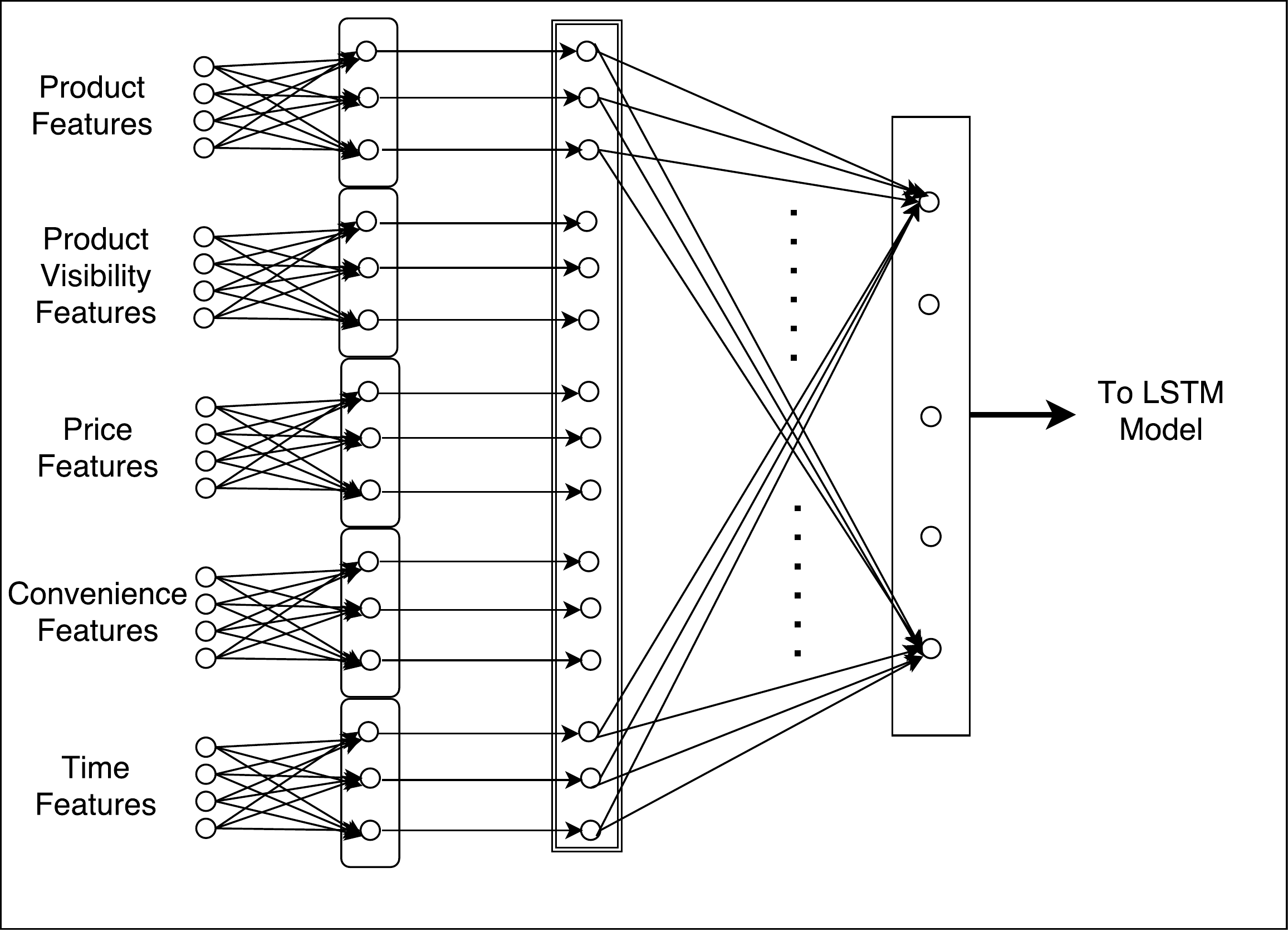}
  \caption{The architecture of the Multi-Layer Perceptron network that models the Associative Layer. The causal or associative features are grouped into five buckets as described in Table~\ref{tab:features}. Embeddings are learned to represent the categorical and 1-Hot features. A fully-connected layer is used to compress the concatenated embeddings into a dense 50-dimensional space. These embeddings are then fed into an LSTM that models the time-series information.}
  \label{fig:dim2}
\end{center}
\end{figure}

\begin{figure}[t]
\begin{center}
  \includegraphics[width=80mm, scale=0.8]{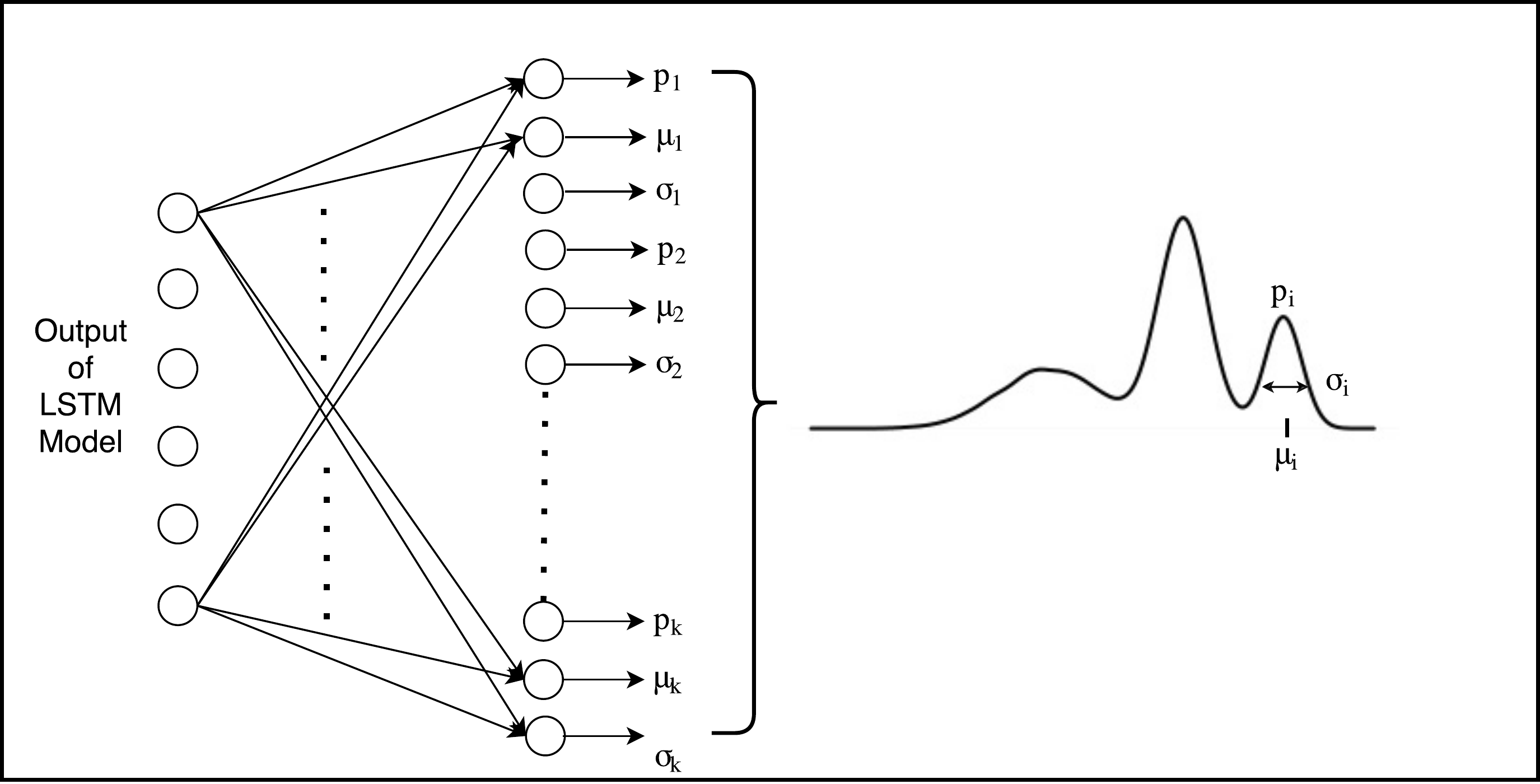}
  \caption{Representation of the Mixture Density Network (MDN) layer which is also the output layer of AR-MDN. The MDN receives as input,  the output from the Recurrent LSTM layer. The MDN learns a fully-connected layer that predicts a mixture of $k$ Gaussians, along with the probability of choosing each Gaussian. This effectively models a multi-modal output space, where the predicted value could be sampled from one of several Gaussians. }
  \label{fig:dim3}
\end{center}
\end{figure}

\subsubsection*{The Associative Layer}
The primary motivation behind this layer was to treat the associative variables that drive demand. Thus, the associative layer is functionally analogous to the regression-based models discussed in Section~\ref{sec:litreview}.
The  model is executed independently for each time point, the output of which, $$\textbf{ff}_{i,t}=\text{MLP}(\vx_{i,t},\Theta^{'})$$ where $\Theta^{'}$ are the parameters of the MLP.

A representative architecture of the feed-forward neural network that models the Associative layer is presented in Figure \ref{fig:dim2}. The features are first bucketed into five types as illustrated in Figure 2 and Table 1.  The features under the ``Price" and ``Time" buckets are all continuous and are simply normalized to zero mean and unit variance. The rest of the feature buckets are composed of  categorical and 1-Hot features which are embedded into a suitable feature space of size $30$. Following this, the entire set of features are concatenated and embedded into a lower-dimensional space of $50$. The final embedding was executed using a single FC-layer with an Exponential Linear Unit (ELU) activation, which was shown to have better numerical stability than the popular ReLU~\cite{clevert15} .

\begin{figure*}[ht]
\centering
   \fbox{\includegraphics[height=50mm, width=80mm]{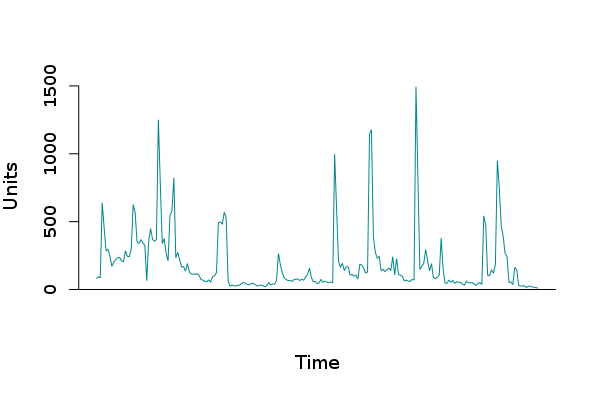}}
    \fbox{\includegraphics[height=50mm, width=80mm]{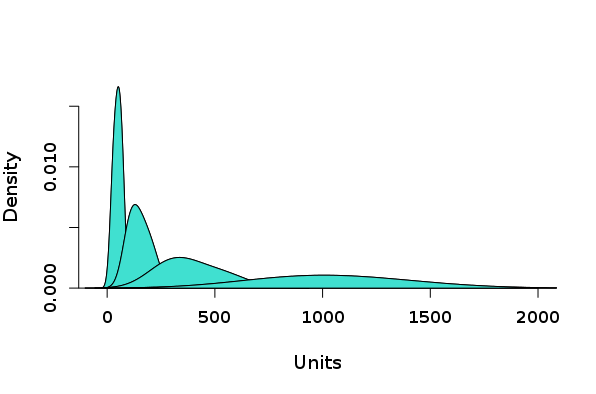}}
    \begin{tabular}{p{8cm}p{8cm}}
    \centering (a) & \centering (b) \\
    \end{tabular}
	\caption{(a) An example time-series of the demand for a particular SKU. One can observe the large number of peaks and valleys in the time-series. It is obvious that a simple time-series model cannot explain this data, and one needs to model the associative factors that contribute to the variance. (b) The Gaussian mixture density of the time-series in (a). It is obvious that the values of the time series are best represented by a multi-modal Gaussian which motivates our use of an MDN layer as the output of our model.}    
    \label{fig:mdn}
\end{figure*}

\subsubsection*{The Recurrent Layer}

As mentioned earlier, the weakness of  Regression based models is that they do not explicitly model the time-series data. The model needs to capture  both the short-term and long-term signals that are present in a time-series data. To achieve this, the output of the Associative layer is fed into a recurrent neural network (RNN).  A popular type of RNN  is LSTM that associates each cell with an input gate, an output gate and a forget gate. The memory unit of an LSTM cell stores the hidden representation of the sequence of input data seen till a given time instant. The LSTM cell learns to either update or reset the memory unit depending on the given sequence of values. The LSTM model integrates the output of the associative model along with {\em all} previous demand and inputs along the time-series as:
\begin{equation}
\vh_{i:t} = L(\{y_{i, 1},\ldots, y_{i, t-1}\}, \{\textbf{ff}_{i, 1},\ldots, \textbf{ff}_{i, t}\}, {\Theta^{''}}).
\end{equation}
where $\Theta^{''}$ are the parameters at this stage.  Internally, the variable length history is stored in LSTMs as state $\vs_{i:t}$ at each point $t$.
At each step, the update in terms of the previous $y$ and current input $\textbf{ff}$ then becomes:
\begin{equation}
\vh_{i:t}, ~\vs_{i:t} = \text{LSTM}({y}_{i,t-1}, \textbf{ff}_{i,t}, \vs_{i:t-1}, {\Theta^{''}}).
\end{equation}
In the above $\vh_{i:t}$ is the vector that is fed to the next output layer and $\vs_{i:t}$ is the state that is stored in this layer and used for computing the outputs at the next time step.  We can view $\vs_{i:t}$ as the auto-regressive features that has been automatically computed by the deep network.   Hence, by combining the ability of the MLP to factor in associative features with the explicit treatment of the sequential nature of a time-series via an LSTM, AR-MDN is able to harness the signals embedded within the data better than either of the layers alone.


\subsubsection*{The MDN Layer}
The output of a typical Neural Network minimizes a mean-squared or a softmax loss function. However, neither of these losses are suitable towards modeling the variation in the demand curve, as can be seen in the example of Figure~\ref{fig:mdn}(a). Minimizing the sum of squared error yields an output that follows a single Gaussian distribution, which would not be powerful enough to model the variation in the output space.  Softmax, on the other hand, is a generalization of the sigmoid function that models a Multinoulli output distribution. Softmax is more suited to classification tasks rather than the regression tasks that we are interested in. 

Instead, the central hypothesis used for modeling the output distribution in our case, is that the regression being performed is multi-modal. As see in Figure~\ref{fig:mdn}(b),  the density plot for the time-series data in Figure~\ref{fig:mdn}(a) is clearly multi-modal, which results in multiple values of $y_i$ for similar values of $x_i$. 

Since Gaussian mixtures is the natural representation for this kind of problems, we use a mixture density network (MDN) as an output layer~\cite{bishop94,zen14}.
The MDN hypothesis is intuitively robust considering the many external factors that are not accounted for in the model.
Therefore, considering $K$ Gaussian mixtures, the conditional distribution can be equated as 
\begin{equation}
P(y_{i,t}|\vh_{i,t})=p_1 N(\mu_1(x),\sigma_1^2(x))+ \ldots +p_K N(\mu_K(x),\sigma_K^2(x))
\end{equation}
where $p_k$,$\mu_k$ and $\sigma_k$ are the probability, mean and standard deviation of the $k^{th}$ Gaussian component respectively.
 These parameters of the GMM are computed from $\vh_{i,t}$ output from the recurrent layer as follows: (We use $h$ instead of  $\vh_{i,t}$ to reduce clutter.)
\begin{equation}
p_k=\frac{exp(z_k^{(p)}h)}{\sum_{l=1}^{K}z_l^{(p)}h}
\label{eqn:softmax}
\end{equation}
\begin{equation}
\sigma_k=exp(z_k^{(\sigma)}h)
\end{equation}
\begin{equation}
\mu_k=z_k^{(\mu)}h
\end{equation}
where $z_k^{(p)}$, $z_k^{(\sigma)}$ and $z_k^{(\mu)}$ are the learned parameters  of the output layer corresponding to the mixture probability, variance and mean respectively. The use of the softmax function in Equation~\ref{eqn:softmax} ensures that the values of $p_k$ lie between 0 and 1 and sum to 1 as is required for probabilities.


Our model distribution can be factorized as a product of likelihood factors over time. The conditional distribution of the network is therefore trying to model is a function of the AR-MDN output $h_{i,t}$, given in Equation 3, and in likelihood terms can be expressed as
\begin{equation}
\begin{split}
P_\Theta (\textbf{y}_{i,t:T}|\textbf{y}_{i,1:t-1},\textbf{x}_{i,1:T})&=\Pi_{t=t}^T P_\Theta(y_{i,t-1}|\textbf{y}_{i,1:t-1},\textbf{x}_{i,1:T})\\ &=
\Pi_{t=t}^T  P({y}_{i,t}|\textbf{h}_{i,t}, \theta) 
\end{split}
\end{equation}
Thus, from the last layer, we get a well-defined joint probability distribution over the demand values for all times periods in the forecast horizon.

\begin{table*}[ht]
\centering
\begin{tabular}{||c c c c c c ||}
  \hline
  \multicolumn{6}{||c||}{\textbf{Test set wMAPE: AR-MDN}} \\
  \hline
  \textbf {Test Window} & \textbf{Week 1} & \textbf{Week 2} & \textbf{Week 3} & \textbf{Week4} & \textbf{Overall} \\
  \hline
  Weeks 9-12 & 22.90 & 27.6 & 31.06* & 36.06 & 29.58 \\
  Weeks 13-16 & 31.98 & 35.71* & 35.97 & 37.02 & 35.17 \\
  Weeks 15-18 & 30.86 & 31.90 & 32.74 & 37.01* & 33.13 \\
  \hline
  Average & 28.58 & 31.74  & 33.26 & 36.93 & 32.63 \\
  \hline
  \multicolumn{6}{||c||}{\textbf{Test set wMAPE: Boosted Cubist}} \\
  \hline
  Weeks 9-12 & 28.51 & 33.21 & 38.63* & 43.26 & 35.63 \\
  Weeks 13-16 & 35.33 & 39.87* & 44.12 & 46.17 & 41.70 \\
  Weeks 15-18 & 32.76 & 36.42 & 41.17 & 43.73* & 39.98 \\
  \hline
  Average & 32.30 & 35.50  & 41.31 & 44.39 & 39.06  \\
  \hline
\end{tabular}
\caption{Comparison of weekly wMAPE for AR-MDN vs Boosted Cubist across four weeks for three different test windows. AR-MDN clearly out-performs the Cubist model in all test cases. The performance gap widens as the prediction horizon increases, with AR-MDN being more stable over time. The weeks that contain a major sale event are marked with an asterix, please see Section~\ref{sec:saleevents} for discussion. Note that with the wMAPE measure, lesser is better.}
\label{table:1}
\end{table*}

\subsection{Training and Implementation}
 \label{sec:train}

Given the scale and diversity of products sold in a large eRetail store like Flipkart, a single set of model parameters $\theta$ over all products in the store is not appropriate, given the large fluctuations in the sale patterns of products within the store.
The other extreme of training a separate model for each SKU raises challenges of data sparsity, particularly for new and emerging products.  We next discuss our strategy of balancing the conflicting challenges of product heterogeneity and data sparsity in training a learned model.

\paragraph*{Trading data heterogeneity and sparsity during training} 
We first trained a joint model, combining all SKUs across all verticals.
A ``vertical'' is a set of products that belong to the same category. Example verticals could be Laptops, Headphones, etc. We then fine\-tuned this model per vertical, training with only the SKUs within the vertical.  To augment the relative sparsity of data at the SKU level, all SKU's belonging to a common vertical (which are relatively homogeneous) are trained together.   

Along the geography dimension, we handle hierarchies a bit differently.  We train first total demand at a  national level which are in a subsequent step distributed to the FC level according to historical ratios. This is necessitated by the fact that data at the FC level is quite sparse and the causal factors like offers and additional visibility is applied only at the national level and not implemented at regional or FC levels.
Historical sales data is available at regional granularity. We remove the outliers from the data and based on 8 weeks of history, compute the percentage sales contribution of each region with respect to overall national demand. Using these ratios, we divide the national level forecast into regional level sales demand. 


\paragraph*{Loss function}
Since the last layer outputs a probability distribution over the demand values, we can resort to the maximum likelihood principle for training the model parameters.  
Given the parameter definition in the output layer, 
the loss function of the model is then given as
\begin{equation}
-\frac{1}{N}\sum_{i=1}^N\sum_{t=1}^{t_0}\log\sum_{k=1}^{K}\frac{p_{i,t,k}}{\sqrt{2\pi\sigma_{i,t,k}^2}}\exp(-\frac{(y_{i,t}-\mu_{i,t,k})^2}{2\sigma_{i,t,k}^2})
\label{eqn:mdn}
\end{equation}
We evaluate this loss for each product's national level sales at each point in the training week.  Training is done using standard stochastic gradient steps.

MDN loss is known to be highly unstable, essentially because of the division by sigma in Equation~\ref{eqn:mdn}~\cite{goodfellow16}. Since $\sigma$ is a predicted value of the Neural Network, it can become arbitrarily small. This can lead to exploding activations, and consequently exploding gradients. To prevent this we employed activation clipping. The value of sigma was clipped to the range $[10^{-5}, 10^{10}]$ . This allowed us to successfully train the model. The number of mixtures were empirically chosen to be 10 by hyper-parameter tuning.


\section{Experimental evaluation}
\label{sec:expts}
In this section we show a detailed evaluation of different forecasting models on Flipkart's operational data.  We compare our best effort traditional model (Boosted Cubist) with a modern deep learning based model (AR-MDN).  We are aware of no earlier comparisons of these two families of models on large-scale industrial setup such as ours. Next, we justify various design choices in the AR-MDN model by showing comparisons with existing simpler variants.
Our experiments are particularly significant because
the rich diversity of our feature space, stresses the limits of state-of-the-art approaches in a way that smaller, public domain datasets cannot.

\paragraph*{Evaluation metric}
Since the downstream business use-case of the forecasting pipeline is buying and replenishment of the inventory, forecasting errors for SKUs which sell a larger number of units have larger negative impact than equivalent error on SKUs which sell lesser units. Hence, the error metric chosen was weighted mean absolute percentage error (wMAPE) at time $t$ and is given by the equation:
\begin{equation}
wMAPE_t=\frac{\sum_{i=1}^{s}|{Y}_{i,t}-y_{i,t}|}{\sum_{i=1}^{s}{Y}_{i,t}}
\end{equation}
where $s$ is the total number of SKUs, ${Y}_{i,t}$ is the actual number of units sold at time $t$ and $y_{i,t}$ is the model forecast at time $t$. 

However, in an industrial setup, we are interested in not just aggregated errors but also need to dissect the aggregates along various facets of product and time.

\paragraph*{Training and Test Data}
The data used for our experiments is collected from October 2015 to March 2017. We evaluate the model on three distinct  test periods of a duration of 4 weeks each: weeks 9-12, 13-16, and 15-18 of 2017. For each test data set, the remaining data (from October 2015 up to the test period) is used as training data. For each experiment, for example week 15-16, demands for all weeks before week 14 of 2017 are fed to the AR-MDN, and predictions obtained for week 1, 2, 3, and 4 after that are compared with actual demand at weeks 15, 16, 17, 18 respectively. 

\paragraph*{Implementation}
The AR-MDN was implemented using TensorFlow, and was trained on a single Titan X GPU with 3072 cores and 12 GB RAM. We used Adam Optimizer and an exponential learning decay policy, decaying learning rate by 0.96 every 1000 iterations. Training the combined model took around 1-2 days, and per vertical fine-tuning took an additional 4-5 days. The total training process, therefore, took around a week. This time includes evaluation that was carried out after each epoch. Running the trained model takes ~100ms/SKU on a 4-core Intel Xeon CPU machine.
We also used dropout of 0.5 after the MLP, and before the LSTM input cells, to prevent overfitting. We used mini batch gradient descent, sampling 512 SKUs for each mini\-batch randomly. The sequence length was chosen to be the maximum sequence length in the mini batch. Zero padding had to be done for rows, which had less data than sequence length of the batch. We later masked the loss for these zero padded data points. LSTM was dynamically unrolled for this sequence length.

\paragraph*{Methods compared}
In Section~\ref{sec:litreview} we discussed a number of methods for demand forecasting. We present comparisons with existing methods in two stages: we first select the best of the breed non-deep learning based method and compare with that.  Next, we compare with various architectures among deep learning models.

\begin{figure*}
\begin{center}
\fbox{\includegraphics[width=0.9\hsize]{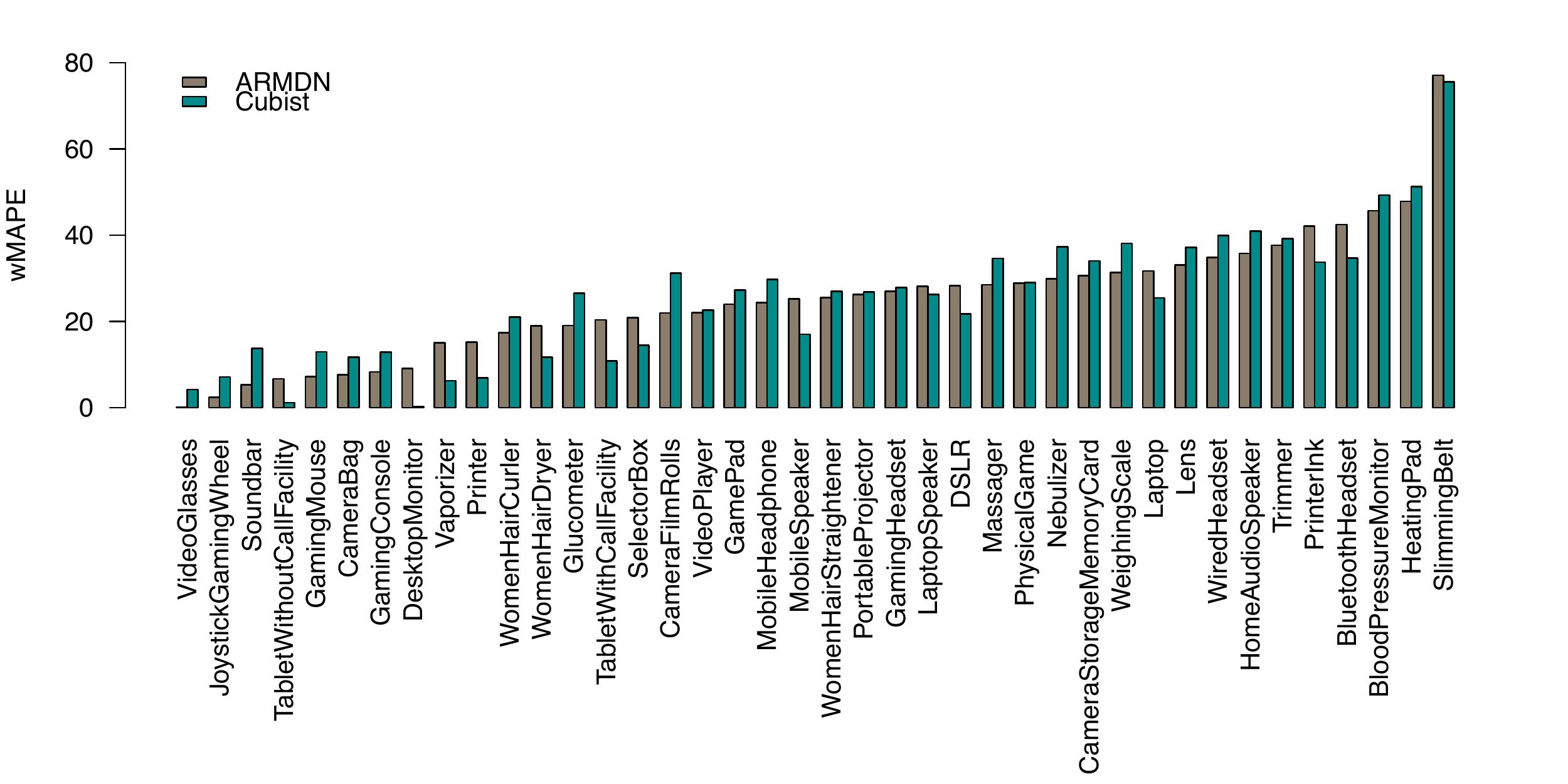}}
  \caption{Comparison of AR-MDN vs Cubist across at vertical level, for a sample set of categories in our test dataset. The AR-MDN model performs better than Cubist for 27 categories out of 40.  Insights from this experiment are discussed in Section~\ref{sec:vertresults}.}
  \label{fig:dim6}
\end{center}
\end{figure*}

\subsection{Traditional Machine Learning Alternatives}
We eliminated pure time-series based models like ARIMA early on because of their inability to capture the effect of the large number of exogeneous factors causing large swings in outputs. Among the regression methods,  the best performing method was a committee of Cubist~\cite{quinlan92} models. Cubist\cite{kuhn13} is an extension of Quinlan's M5 model~\cite{quinlan92,quinlan93,quinlan93a} model that skillfully combines three learning structures: decision trees at the top-level to capture non-linearity,  regression models at the leaves to encourage smooth predictions, and finally $k$ nearest neighbor based refinement.  We trained a committee of $M$ Cubist models via Boosting.  The values of $k$ and $M$ were fixed at 9 and 50 via time-slice validation.  The Cubist models used the same set of features as AR-MDN as outlined in Table~\ref{tab:features}.  However, since Cubist is a scalar regression model, and not a time series model, we created a set of autoregressive features and temporal features along time.  Specifically we used auto-regressive features like mean sale of the previous week and mean sale of the previous month; and temporal features like week of year and time since first sale/launch. 

\begin{figure}[ht]
\begin{center}
 \fbox{\includegraphics[width=80mm, scale=0.8]{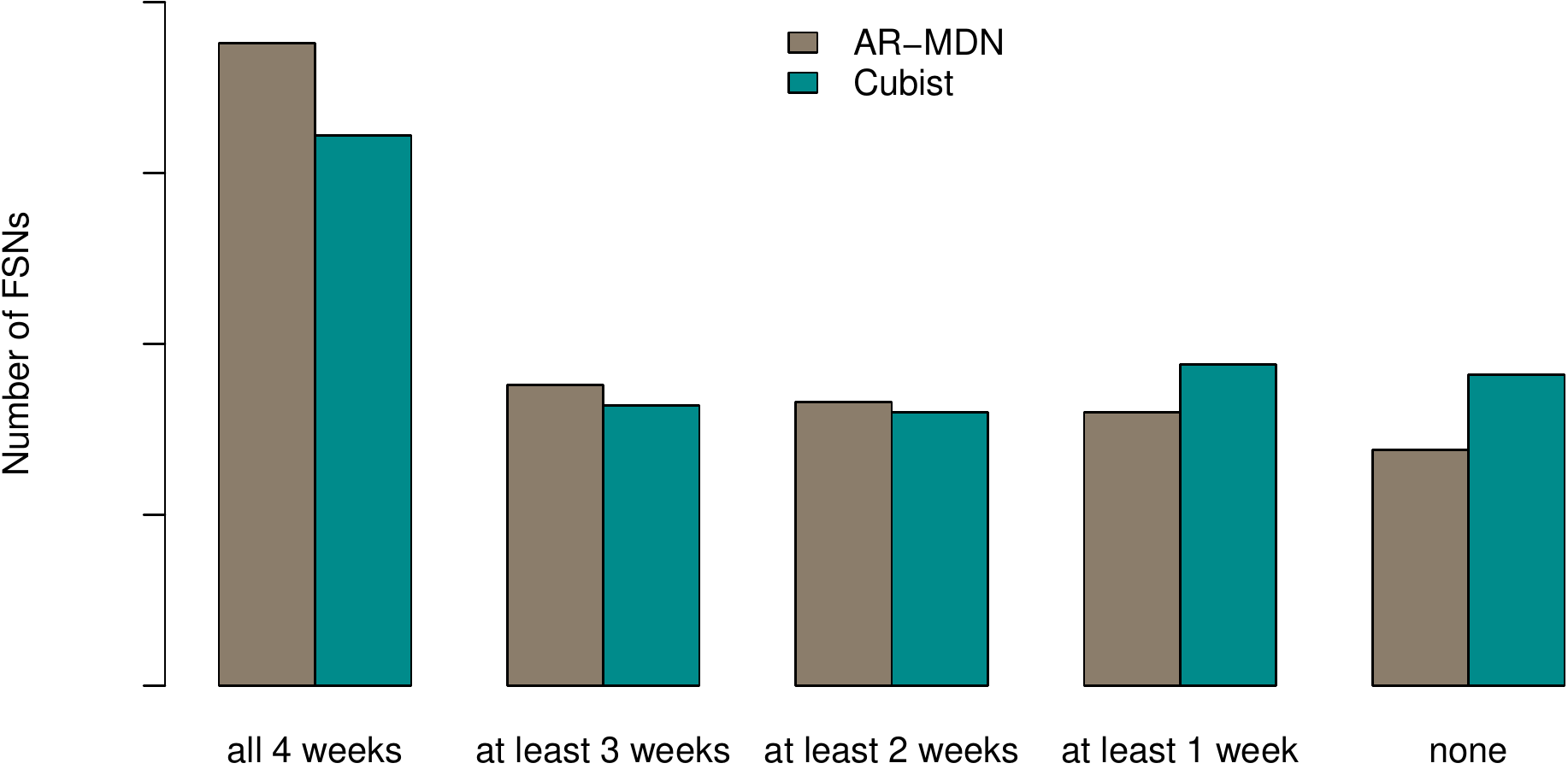}}
  \caption{Towards deploying the model, a prediction of wMAPE less than 30\% is considered reliable enough to automate the replenishment process. Here we show the percentage of SKUs that are ``actionable''  across the various weekly time horizons, compared across the AR-MDN solution and the baseline Cubist. The AR-MDN predictions are clearly  actionable across a larger set of SKUs. }
  \label{fig:skuweeks}
\end{center}
\end{figure}

Table \ref{table:1} shows the comparison of the results obtained by the AR-MDN model and the Boosted-Cubist model  across four weeks for three different test windows.
Overall, AR-MDN has a wMAPE between $22.90\%-37.02\%$ across the three sets of experiments whereas Boosted Cubist has a higher error of $28.51\%-46.17\%$.  
The AR-MDN model consistently outperforms the Cubist model both of which greatly outperform the ARIMA model (data not shown). 

Another interesting trend is the difference in error between the Cubist model and AR-MDN model increases as the forecasting horizon increases. For the 1-week-out forecast the difference in wMAPE is in the range of $2-5\%$ which increases to $7-9\%$ for the 4-weeks-out forecasts.

The performance is also slightly worse during weeks where a multi-day large sale event occurred (weeks 11, 14, and 18) as opposed to the ``business as usual" weeks i.e. weeks which did not feature large sale events. We will discuss further on such event weeks in Section~\ref{sec:saleevents}.

\subsubsection{Actionable Performance}

Another important parameter for success called ``hits" can be defined as the number of SKUs for which the forecasts are actionable i.e. has average percentage error less than $30\%$ across all three testing windows. The cutoff of $30\%$ was chosen based on business inputs.  For AR-MDN $69\%$/$61\%$of the SKUs are ``hits'' for 1-week-out and 4-weeks-out forecasts respectively while the corresponding numbers for the Cubist model are $64
\%$/$52\%$. Hence, a similar trend holds that the AR-MDN outperforms the Cubist and difference is more notable when the prediction time horizon is larger.

Finally, we measure the quality of the models at the SKU level. In Figure \ref{fig:skuweeks} we show the number of SKUs which have lesser than cutoff error across weeks. Based on Figure \ref{fig:skuweeks}, it can be concluded that almost $70\%$ of the SKUs are actionable and the replenishment of these can be completely automated based on the described models.

\begin{figure*}[ht]
\begin{center}
 \fbox{\includegraphics[width=\hsize]{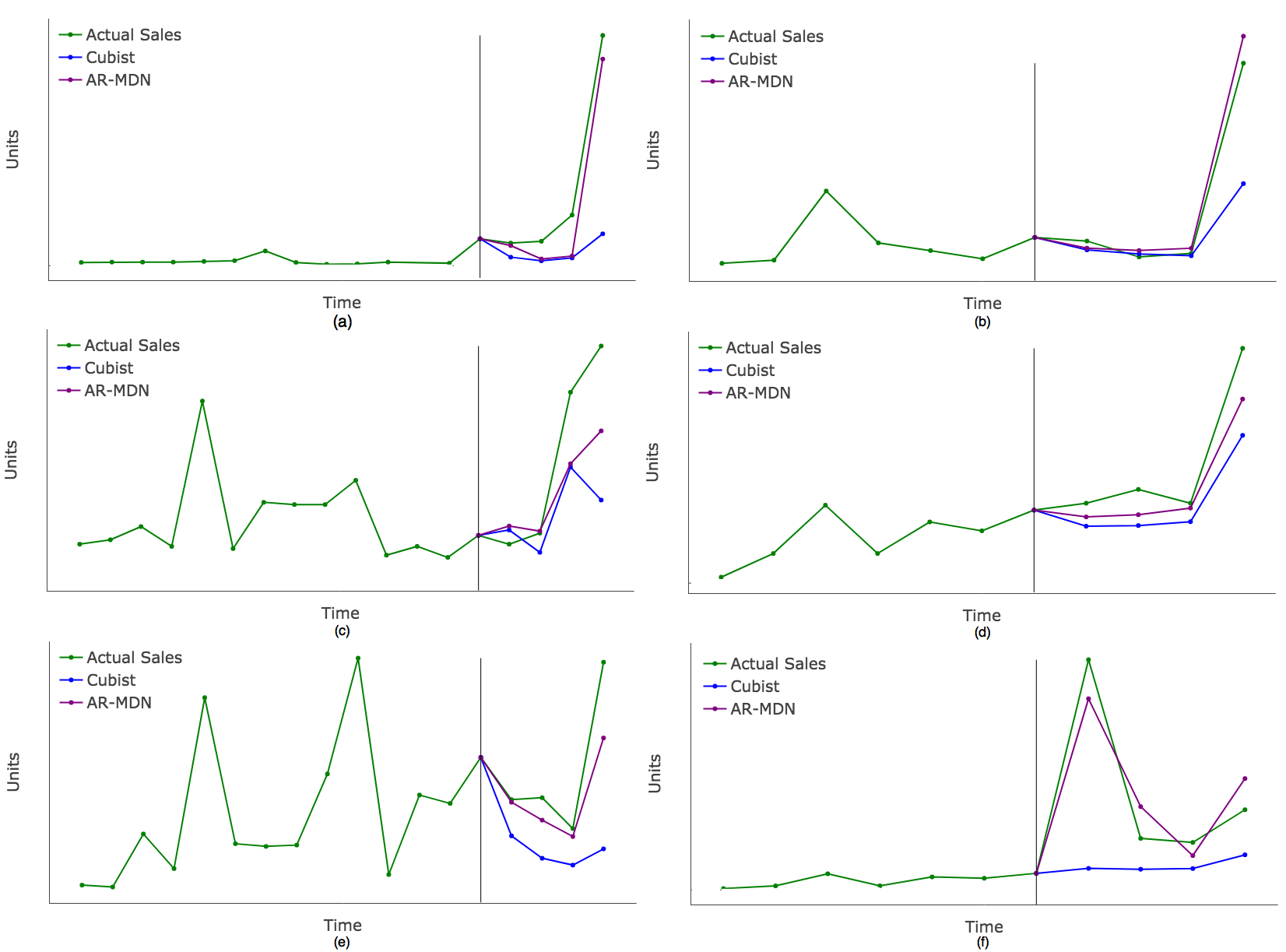}}
  \caption{The performance of the AR-MDN model compared with the actual sale values and with the baseline for an example set of SKUs. The training data is to the left of the vertical line, and the test period is to the right. For the SKU shown in (a) to (e), the $4^{th}$ week contained a major sale event, hence the spike in number of units sold. For the SKU in (f) the sale event was in the $2^{nd}$ week of testing. See further discussion in Section~\ref{sec:skudisc}.}
  \label{fig:skuexamples}
\end{center}
\end{figure*}

\begin{figure}[t]
\begin{center}
 \fbox{\includegraphics[width=0.9\hsize]{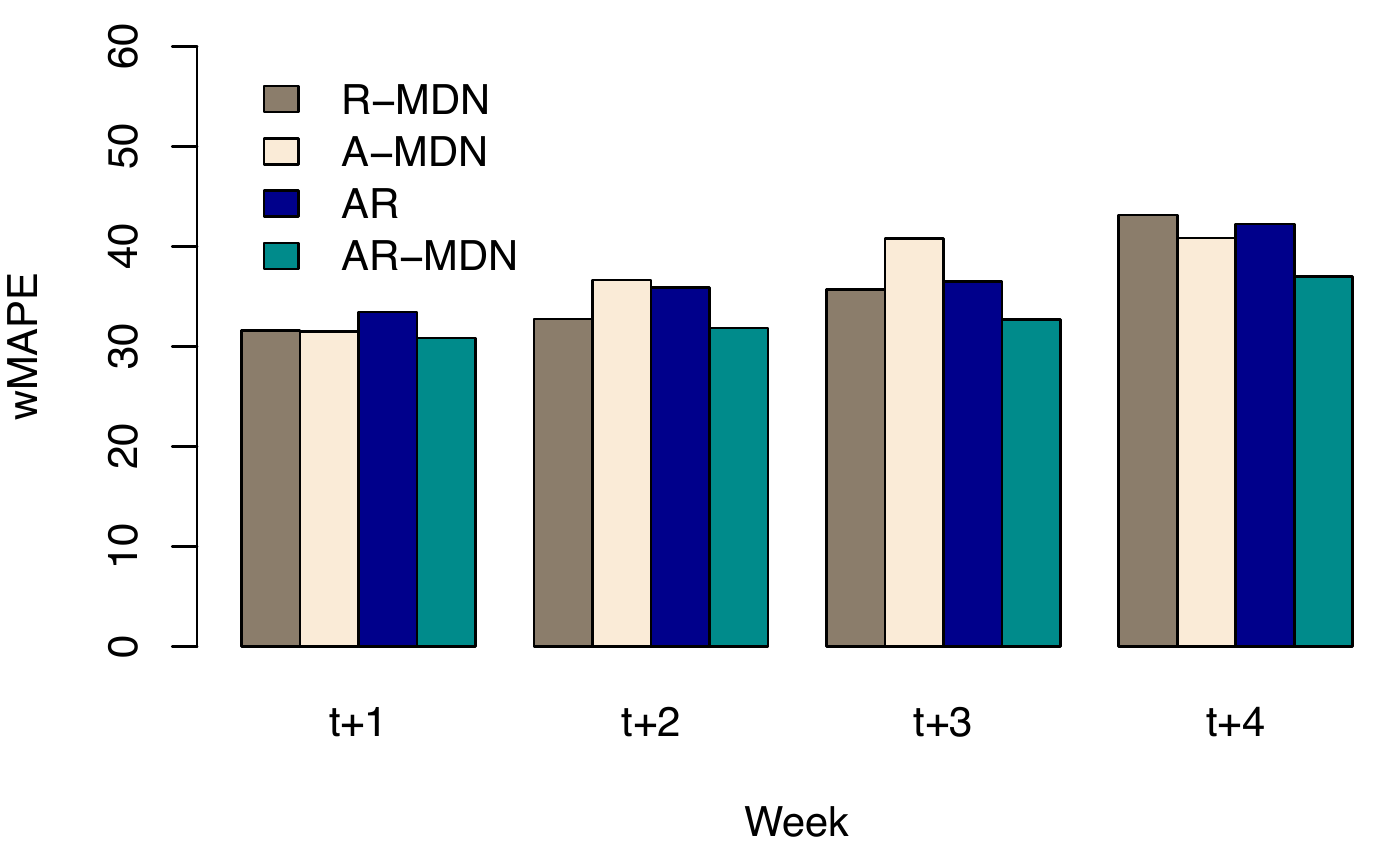}}
  \caption{Comparison of AR-MDN with alternative Neural-Network model architectures at the SKU level. It is clear that the complete AR-MDN architecture outperforms the ablated models. Note that the 4-weeks-out test case ($t+4$) contained a major sale event, hence the particularly large error.}
  \label{fig:ablation}
\end{center}
\end{figure}

\begin{figure}[t]
\begin{center}
\fbox{\includegraphics[width=0.95\hsize]{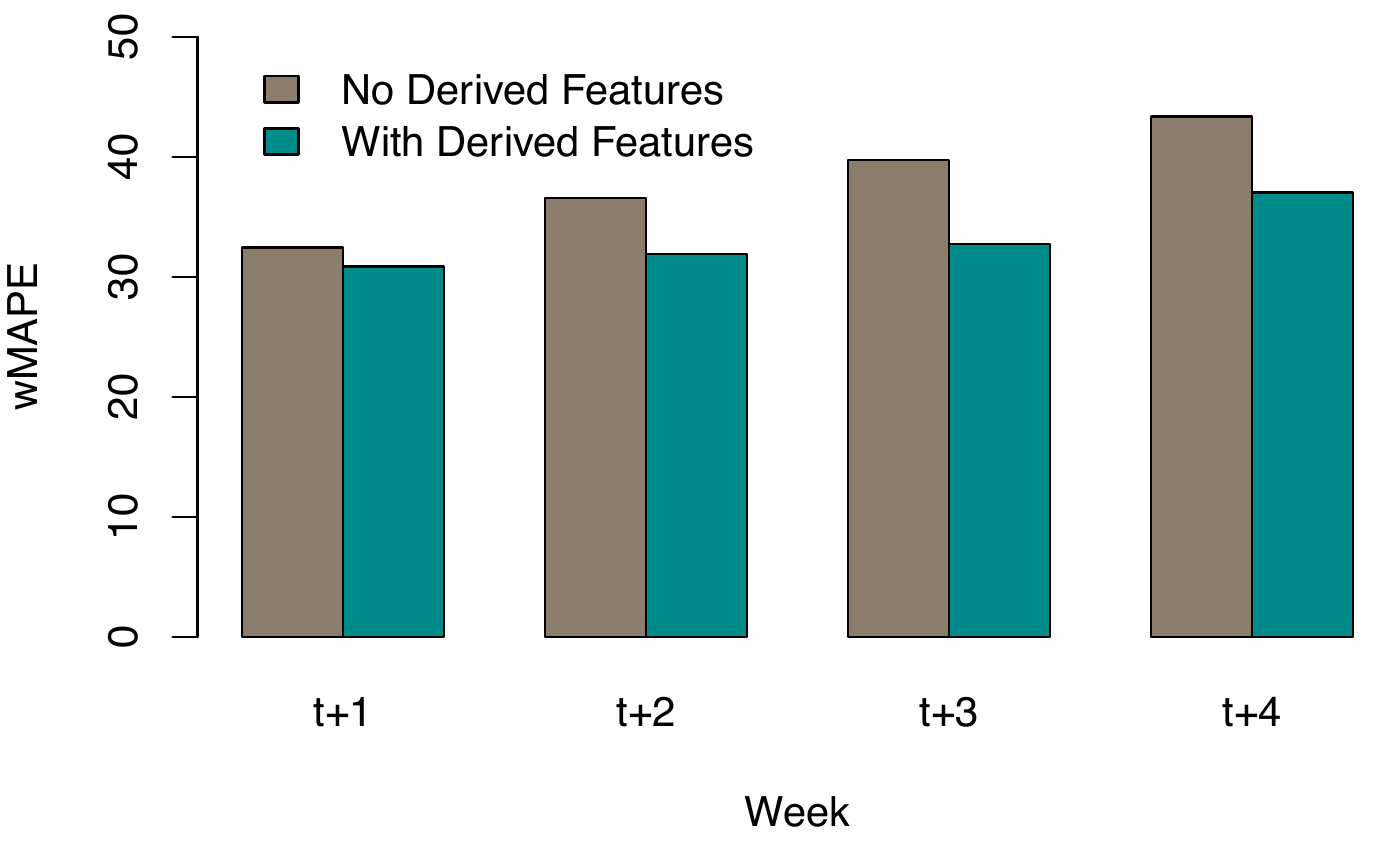}}
  \caption{Comparison of prediction error with and without the use of derived features. Though the derived features are obtained from the raw features, the model cannot explicitly capture these novel signals. The model benefits from including several ``expert-designed'' features that could lessen the complexity of the machine learning.}
  \label{fig:features}
\end{center}
\end{figure}

\subsubsection{Vertical-Level Results}
\label{sec:vertresults}
While the models are quite robust at the SKU level, it was imperative to understand whether the models performed better across certain verticals (products belonging to the same category), and how consistent the performance was across the product classes. In Figure~\ref{fig:dim6} we show the performance of both models at a vertical level, where it becomes clear that performance is fairly consistent for both models.  
However, out of the 40 verticals, AR-MDN does better than the Cubist model in 27 out of 40 verticals. Since, verticals are a homogeneous mix of SKUs, and are a ``decision making unit" from business process perspective, this is a significant data point. 

The decision on which of the two models would be adopted as a business process depends on the overall performance of the models on a vertical.  Given that the number of SKUs are numerous it is not practically possible to make the choice of model at an SKU level. Therefore, given the performance of AR-MDN, it would be more widely integrated into the business process as opposed to Cubist. 

\subsubsection{Performance on Representative SKUs}
\label{sec:skudisc}

In Figure \ref{fig:skuexamples} we show the performance of AR-MDN and Cubist on a representative set of 6 SKUs for test window 3. In the first 5 SKUs (Figure \ref{fig:skuexamples} (a)-(e)), the $4^{th}$ time-point or Week 18 contains a large sale event. In general, all SKUs in our dataset show sharp increase in demand during an event week, a trend observed here as well. However, these 6 SKUs represent slightly different classes in terms of sale trend and we discuss them individually below. For confidentiality reasons, we are unable to reveal the actual product corresponding to each SKU.

\begin{description}

\item[(a)] This SKU is representative of products that have very low demand during business-as-usual (BAU) weeks. However,  they show orders-of-magnitude jump in demand during a sale event, in response to the associative factors that drive a large event. AR-MDN clearly outperforms Cubist in picking up this drastic change of demand. While Cubist does pick up the increasing trend, it falters in correctly picking the magnitude of increase. 

\item [(b)] This SKU represents those that generally show moderate but steady sales during a BAU week and then the demand jumps sharply during the event week. Here, while both AR-MDN and Cubist do well in picking the trend during the non-event time points, AR-MDN clearly outperforms Cubist in estimating the magnitude of the demand jump during the event week. 

\item[(c)] Represents SKUs that have fluctuating demand but shows sharp jumps just prior to an event (around the time frame when marketing and advertising campaigns begin). The increasing trend is maintained during the event week. These SKUs typically see ``warm-up'' offers before the actual event, to cater to the interests of additional visitors to the website in response to marketing campaigns. Both Cubist and AR-MDN are successful in modeling the pre-event week rise in demand. However, while AR-MDN correctly detects the continuing increasing tread correctly, Cubist fails to do so. 

\item[(d)] This example is representative of SKUs which are  relatively new products. They show and overall  increasing trend, which is then boosted by a sale event. Both Cubist and AR-MDN are moderately successful, with AR-MDN doing slightly better, in predicting the non-event and event week demand. 

\item[(e)] This SKU is representative of products that display wildly fluctuating demands on a weekly basis  with sharp peaks and deep valleys due to various known and latent associative factors. Event weeks show sharp rises but the magnitude of the rise may have been observed during non-event weeks as well. Cubist fails in detecting the inherent fluctuating trends and forecasts a steady decrease while the AR-MDN does a much better job in modeling the sharp rises and falls.

\item[(f)] This example represents SKUs which generally have low sales during non-event weeks, participates but not strongly during an event week but rather tend to have their own "mini" events. Once again, Cubist fails to pick up the idiosyncratic nature of demand for this SKU while AR-MDN forecasts  fairly accurately.
\end{description}

While these 6 SKUs are only a snapshot of the variety of demand patterns in our dataset, they clearly highlight the advantage of AR-MDN across a large cluster of products.

\vspace{1cm}


\subsection{Comparison with Alternative Neural Model Architectures}

The AR-MDN model is composed of three modular sub-parts; i) MLP as the associative layer, ii) LSTM as the recurrent layer, and iii) MDN as the output layer. As stated above, the rationale for this modular architecture was that the MLP is useful for the associative features affecting demand, the LSTM handles the sequential aspects of demand while the MDN is necessary to handle the multi-modality of the output space.
The question naturally arises whether, the hypothesis is in fact correct and whether all three subparts are integral to the performance of the machine. Further, how much is performance affected without a particular sub-part.

To quantify the above a series of experiments were performed by removing one component at a time from the AR-MDN model. Therefore we developed a R-MDN (no MLP), A-MDN (no LSTM), AR (single Gaussian)  and compared against the complete AR-MDN model.   Existing deep learning models for time series predictions are based purely on a LSTM with neither the mixture of Gaussian or the multi-layer associative model~\cite{flunkert17}.

Figure~\ref{fig:ablation} compares the error of AR-MDN with these alternative architectures for the weeks 15-18 test window (the same trends holds for other test windows as well). From these experiments we can make the following observations:
\begin{itemize}
\item We clearly see that the complete AR-MDN machine outperforms all its variants by varying degrees.
\item A model without the associative layer (R-MDN) has between 1\% to 7\% higher error than full AR-MDN. The associative layer has a greater role in forecasts further out in the future than in the next week.
\item The recurrent layer also has an important role in reducing error and its removal causes error to increase from 0.5\% to 8\%.
\item The mixture of ten Gaussians in the output layer is significantly better than a single Gaussian across all four forecast horizons. 
\end{itemize}

Overall, the experiments prove that each of the layers of the hybrid AR-MDN architecture is integral to the performance and the layers working in tandem unleash the full modeling prowess of the machine.

\subsection{Effect of Engineered Features}
\label{sec:derived}

Deep learning promises to remove the need for feature engineering~\cite{LeCunBH15}, and on many speech and image processing tasks that promise has been largely met~\cite{hintonspeech, razavian14}. In these experiments we show that for our environment of highly heterogeneous features and products, features engineered by domain experts continue to be extremely useful.

In Section~\ref{sec:features} we described a number of derived functions.  These features are used to capture inherent and easily derivable properties in the time series and price and discounting related observations. 
%
%
In Figure~\ref{fig:features} we plot error with and without our hand engineered features.  We find that these features had a significant impact in reducing error.  These features are particularly useful for predictions further into the future where the recurrent features tend to be less useful.

 
\subsection{Performance During ``Event'' Weeks}
\label{sec:saleevents}
Flipkart tends to hold small to large scale events frequently, spanning multiple days, which are characterized by larger than usual discounts, aggressive marketing campaigns and strong word-of-mouth advertising. The ``event'' weeks also tend to have higher number of unknowns or latent variables like regional festivals, matching competitor events, launch of a popular and coveted product, etc. Importantly, these weeks are also a time for maximizing opportunity (high sale numbers) and when the supply-chain is most stressed due to the higher volumes. Business planning (for example, inventory buying) for these events also happen earlier than usual, generally 2-4 weeks in advance. Therefore, error in forecasts (which need to be more than 1-week-out) during the event weeks, disproportionately affects business critical metrics, as compared to "business-as-usual" weeks. In our test windows, weeks 11 (3-weeks-out in test window 1), 14 (2-weeks-out in test window 2), and 18 (4-weeks-out in test window 3) contain such events. 

In the comparison with Cubist in Table~\ref{table:1}, it is amply clear that AR-MDN is the better of the two models for all 3 weeks that contain a major sales event. The average wMAPE for these 3 weeks are $34.59\%$ and $40.74\%$ for AR-MDN  and Cubist, respectively. Given the volume of sales that are observed on these weeks, this difference in performance alone is sufficient motivation for the  production scale use of AR-MDN. 

In the comparison with alternative neural architectures, the insights obtained are even more interesting and clearly highlights the potency of the hybrid architecture deployed in AR-MDN. The snapshot provided in Figure \ref{fig:ablation} is for weeks 15-18 window of the test set.  Here  the event week is 4-weeks-out and represented by the $t+4$ time point in the Figure. R-MDN is the worst performing of the alternatives, and all the other models which do not omit the associative layer beats the vanilla LSTM model. This reinforces the intuitive belief that  the effect of the associative factors are largest during the event weeks. Another interesting observation is that though both AR and A-MDN retain the associative layer, the models with MDN outperforms the one with the single Gaussian. As mentioned above, the event weeks are also peppered with latent variables and the noise tolerance brought about by the MDN has clear advantages. The complete AR-MDN machine significantly outperforms all the other architectures including A-MDN, highlighting that even during weeks where associative factors hold greater importance, the additional temporal features captured by the recurrent layer adds value. The same observations also hold true for weeks 11 and 14 in the other test windows. 
 
\section{Conclusions and Future Directions}
\label{sec:conc}
In this work, we develop a unified ML architecture that can simultaneously model causal factors, time-series trends and multi-modal output spaces. The machine can be trained end-to-end, in reasonable time over a large real-world dataset. Results show that the proposed architecture easily outperforms a state-of-the-art model based on boosted decision trees. 

Deep Learning based solutions have been found to be adept at automatic feature learning for structured data such as Images and Speech. However, for problems over multi-modal data that spans numerical, categorical, binary and time-series modalities, Deep Learning benefits heavily from so-called ``expert-designed'' features. Similar observation was also made previously in ~\cite{youtubereco}. 

One of the key aspects to evaluate in the future, is the effect of prediction errors on metrics such as logistics cost, SLA adherence and overall customer satisfaction. Modeling and evaluating these aspects, while interesting, is beyond the scope of this work. 

Future work shall focus on the forecasting at different granularities of Product $\times$ Geography $\times$ Time. It would be interesting to examine if the solution we proposed in this work, could apply directly to other granularities; or if it would require non-trivial redesign of the architecture. Similarly interesting exploration would be in the re-use of a learned model at a certain granularity to initialize the  model for a different granularity. 

\balance
\bibliographystyle{abbrv}
\bibliography{references}

\begin{thebibliography}{10}

\bibitem{gradientboosting}
Winning Model Documentation for Rossmann Store Sales at:
  https://kaggle2.blob.core.windows.net/forum-message-attachments/102102/3454/Rossmann\_nr1\_doc.pdf.

\bibitem{APPELHANS2015}
T.~Appelhans, E.~Mwangomo, D.~R. Hardy, A.~Hemp, and T.~Nauss.
\newblock Evaluating machine learning approaches for the interpolation of
  monthly air temperature at {M}t. {K}ilimanjaro, {T}anzania.
\newblock {\em Spatial Statistics}, 14:91 -- 113, 2015.

\bibitem{bjorn95}
V.~Bjorn.
\newblock Multiresolution methods for financial time series prediction.
\newblock {\em Proceedings of the IEEE/IAFE 1995 Conference on Computational
  Intelligence for Financial Engineering}, page~97, 1995.

\bibitem{box64}
G.~Box and D.~Cox.
\newblock An analysis of transformations.
\newblock {\em Journal of Royal Statistical Society. Series B
  (Methodological)}, 26(2):211--252, 1964.

\bibitem{box68}
G.~Box and G.~M. Jenkins.
\newblock Some recent advances in forecasting and control.
\newblock {\em Journal of Royal Statistical Society. Series C (Applied
  Statistics)}, 17(2):91--109, 1968.

\bibitem{busseti12}
E.~Busseti, I.~Osband, and S.~Wong.
\newblock Deep learning for time series modeling.
\newblock {\em Technical Report, Stanford University}, 2012.

\bibitem{clevert15}
D.-A. Clevert, T.~Unterthiner, and S.~Hochreiter.
\newblock {Fast and Accurate Deep Network Learning by Exponential Linear Units
  (ELUs)}.
\newblock {\em CoRR}, abs/1511.07289, 2015.

\bibitem{youtubereco}
P.~Covington, J.~Adams, and E.~Sargin.
\newblock Deep neural networks for youtube recommendations.
\newblock In {\em Proceedings of the 10th ACM Conference on Recommender
  Systems}, RecSys '16, pages 191--198, 2016.

\bibitem{cristi00}
R.~Cristi and M.~Tummula.
\newblock Multirate, multiresolution, recursive kalman filter.
\newblock {\em Signal processing}, 80:1945--1958, 2000.

\bibitem{johnson15}
K.~J. Ferreira, B.~H.~A. Lee, and D.~Simchi-Levi.
\newblock Analytics for and online retailer: Demand forecasting and price
  optimization.
\newblock {\em Manufacturing and Service Operations Management}, 18(1):69--88,
  2015.

\bibitem{flunkert17}
V.~Flunkert, D.~Salinas, and J.~Gasthaus.
\newblock Deep{AR}: Probabilistic forecasting with autoregressive neural
  networks.
\newblock {\em arXiv:1704.04110}, 2017.

\bibitem{goodfellow16}
I.~Goodfellow, Y.~Bengio, and A.~Courville.
\newblock {\em Deep Learning}.
\newblock MIT Press, 2016.
\newblock \url{http://www.deeplearningbook.org}.

\bibitem{hintonspeech}
G.~Hinton, L.~Deng, D.~Yu, G.~E. Dahl, A.~r.~Mohamed, N.~Jaitly, A.~Senior,
  V.~Vanhoucke, P.~Nguyen, T.~N. Sainath, and B.~Kingsbury.
\newblock Deep neural networks for acoustic modeling in speech recognition: The
  shared views of four research groups.
\newblock {\em IEEE Signal Processing Magazine}, 29(6):82--97, 2012.

\bibitem{hyndman08}
R.~Hyndman, A.~Koehler, J.~Ord, and R.~Snyder.
\newblock {\em Forecasting with exponential smoothing: The state space
  approach}.
\newblock Springer, 2008.

\bibitem{kleinberg03}
R.~Kleinberg and T.~Leighton.
\newblock The value of knowing a demand curve: Bounds on regret for online
  posted-price auctions.
\newblock pages 594--605, 2003.

\bibitem{KOURENTZES2013}
N.~Kourentzes.
\newblock Intermittent demand forecasts with neural networks.
\newblock {\em International Journal of Production Economics}, 143(1):198 --
  206, 2013.

\bibitem{kuhn13}
M.~Kuhn, S.~Weston, C.~Keefer, and N.~Woulton.
\newblock Cubist: Rule and instance based regression modeling.
\newblock {\em R Package Version 0.0.13, CRAN}, 2013.

\bibitem{Kuremoto2014}
T.~Kuremoto, S.~Kimura, K.~Kobayashi, and M.~Obayashi.
\newblock Time series forecasting using a deep belief network with restricted
  {B}oltzmann machines.
\newblock {\em Neurocomputing}, 137:47 -- 56, 2014.

\bibitem{larson01}
P.~D. Larson, D.~Simchi-Levi, P.~Kaminsky, and E.~Simchi-Levi.
\newblock Designing and manging the supply chain.
\newblock {\em Journal of Business Logistics}, 22(1):259--261, 2001.

\bibitem{LeCunBH15}
Y.~LeCun, Y.~Bengio, and G.~E. Hinton.
\newblock Deep learning.
\newblock {\em Nature}, 521(7553):436--444, 2015.

\bibitem{bishop94}
C.~M.~Bishop.
\newblock Mixture density networks.
\newblock 01 1994.

\bibitem{meyer16}
H.~Meyer, M.~Katurji, T.~Appelhans, M.~Muller, T.~Nauss, P.~Roudier, and
  P.~Zawar-Reza.
\newblock Mapping daily air temperature for {A}ntarctica based on {MODIS LST}.
\newblock {\em Remote Sensing}, 8:732, 2016.

\bibitem{mocanu16}
E.~Mocanu, P.~H. Nguyen, M.~Gibescu, E.~M. Larsen, and P.~Pinson.
\newblock Demand forecasting at low aggregation levels using factored
  conditional restricted boltzmann machine.
\newblock {\em 2016 Power Systems Computation Conference (PSCC)}, pages 1--7,
  2016.

\bibitem{moody97}
J.~Moody and W.~Lizhong.
\newblock What is true price? state space models for high frequency fx data.
\newblock {\em Proceedings of the IEEE/IAFE 1997 Conference on Computational
  Intelligence for Financial Engineering}, pages 150--156, 1997.

\bibitem{papagiannnaki03}
K.~Papagiannaki, N.~Taft, Z.~L. Zhang, and C.~Diot.
\newblock Long-term forecasting of internet backbone traffic: observations and
  initial models.
\newblock {\em IEEE Transactions on Neural Networks}, 2:178--1188, 2003.

\bibitem{qiu14}
X.~Qiu, L.~Zhang, Y.~Ren, P.~N. Suganthan, and G.~Amaratunga.
\newblock Ensemble deep learning for regression and time series forecasting.
\newblock {\em 2014 IEEE Symposium on Computational Intelligence in Ensemble
  Learning (CIEL)}, pages 1--6, 2014.

\bibitem{quinlan92}
J.~Quinlan.
\newblock Learning with continuous classes.
\newblock {\em Proceedings of the 5th Australian Joint Conference on Artificial
  Intelligence}, pages 343--348, 1992.

\bibitem{quinlan93a}
J.~Quinlan.
\newblock C4.5:programs for machine learning.
\newblock {\em Morgan Kaufmann Publishers Inc}, pages 236--243, 1993.

\bibitem{quinlan93}
J.~Quinlan.
\newblock Combining instance based and model based learning.
\newblock {\em Proceedings of the 10th International Conference on Machine
  learning}, pages 236--243, 1993.

\bibitem{razavian14}
A.~S. Razavian, H.~Azizpour, J.~Sullivan, and S.~Carlsson.
\newblock Cnn features off-the-shelf: An astounding baseline for recognition.
\newblock In {\em Proceedings of the 2014 IEEE Conference on Computer Vision
  and Pattern Recognition Workshops}, CVPRW '14, pages 512--519, 2014.

\bibitem{renaud02}
O.~Renaud, J.-L. Starck, and F.~Murtagh.
\newblock Wavelet-based forecasting of short and long memory time series.
\newblock {\em Institut d'Economie et Econométrie, Université de Genève},
  2002.

\bibitem{shi15}
X.~Shi, Z.~Chen, H.~Wang, D.-Y. Yeung, W.~kin Wong, and W.~chun Woo.
\newblock Convolutional lstm network: A machine learning approach for
  precipitation nowcasting.
\newblock {\em Proceedings of the 28th International Conference on Neural
  Information Processing Systems (NIPS)}, 1:802--810, 2015.

\bibitem{soltani00}
S.~Soltani, D.~Boichu, P.~Simard, and S.~Canu.
\newblock The long-term memory prediction by multiscale decomposition.
\newblock {\em Signal Processing}, 80:2195--2205, 2000.

\bibitem{stolojescu2010}
C.~Stolojescu, I.~Railean, S.~Moga, P.~Lenca, and A.~Isar.
\newblock A wavelet based prediction method for time series.
\newblock {\em Proceedings of Stochastic Modeling Techniques and Data Analysis
  (SMTDA)}, pages 757--764, 2010.

\bibitem{Torgo1997}
L.~Torgo.
\newblock Functional models for regression tree leaves.
\newblock {\em ICML}, pages 385--393, 1997.

\bibitem{walton08}
J.~Walton.
\newblock Subpixel urban land cover estimation: Comparing cubist, random
  forests and support vector regression.
\newblock {\em Photogram. Eng. Remote Sens.}, 74:1213--1222, 2008.

\bibitem{WANG2016}
B.~Wang, C.~Oldham, and M.~R. Hipsey.
\newblock Comparison of machine learning techniques and variables for
  groundwater dissolved organic nitrogen prediction in an urban area.
\newblock {\em Procedia Engineering}, 154:1176 -- 1184, 2016.

\bibitem{wang02}
X.~Wang and X.~Shan.
\newblock A wavelet based method to predict internet traffic.
\newblock {\em Communications, Circuits and Systems and West Sino Expositions},
  1:690--694, 2002.

\bibitem{zen14}
H.~Zen and A.~Senior.
\newblock Deep mixture density networks for acoustic modeling in statistical
  parametric speech synthesis.
\newblock {\em Proceedings of the IEEE International Conference on Acoustics,
  Speech, and Signal Processing (ICASSP)}, pages 872--3876, 2014.

\bibitem{zhang01}
B.-L. Zhang, R.~Coggins, M.~Jabri, D.~Dersch, and B.Flower.
\newblock Multiresolution forecasting for futures trading using wavelet
  decompositions.
\newblock {\em IEEE Transactions on Neural Networks}, 12(4):765--775, 2001.

\bibitem{zhang16}
H.~Zhang, F.~Zhang, M.~Ye, T.~Che, and G.~Zhang.
\newblock Estimating daily air temperatures over the {T}ibetan {P}lateau by
  dynamically integrating {MODIS LST}.
\newblock {\em Journal of Geophysical Research: Atmospheres},
  121(19):425--11,441, 2016.

\bibitem{zheng99}
G.~Zheng, J.~Starck, J.~Campbell, and F.~Murtagh.
\newblock The wavelet transform for filtering financial data streams.
\newblock {\em Journal of Computational Intelligence in Finance}, 7:18--35,
  1999.

\end{thebibliography}
\end{document}